\documentclass[sigconf]{aamas}

\usepackage{balance} 
\usepackage{multirow}
\usepackage{makecell}
\usepackage{bm}
\setcopyright{none}
\renewcommand\footnotetextcopyrightpermission[1]{}
\setcopyright{ifaamas}
\acmConference[AAMAS '25]{Proc.\@ of the 24th International Conference
on Autonomous Agents and Multiagent Systems (AAMAS 2025)}{May 19 -- 23, 2025}
{Detroit, Michigan, USA}{Y.~Vorobeychik, S.~Das, A.~Nowé  (eds.)}
\copyrightyear{2025}
\acmYear{2025}
\acmDOI{}
\acmPrice{}
\acmISBN{}

\title[AAMAS-2025 Formatting Instructions]{Making Universal Policies Universal}

\author{Niklas Hoepner}
\affiliation{
  \institution{University of Amsterdam}
  \city{Amsterdam}
  \country{Netherlands}}
\email{n.r.hopner@uva.nl}

\author{David Kuric}
\affiliation{
  \institution{University of Amsterdam}
  \city{Amsterdam}
  \country{Netherlands}}
\email{d.kuric@uva.nl}

\author{Herke van Hoof}
\affiliation{
  \institution{University of Amsterdam}
  \city{Amsterdam}
  \country{Netherlands}}
\email{h.c.vanhoof@uva.nl}

\begin{abstract}
The development of a generalist agent capable of solving a wide range of sequential decision-making tasks remains a significant challenge. We address this problem in a cross-agent setup where agents share the same observation space but differ in their action spaces. Our approach builds on the universal policy framework, which decouples policy learning into two stages: a diffusion-based planner that generates observation sequences and an inverse dynamics model that assigns actions to these plans. We propose a method for training the planner on a joint dataset composed of trajectories from all agents. This method offers the benefit of positive transfer by pooling data from different agents, while the primary challenge lies in adapting shared plans to each agent’s unique constraints. We evaluate our approach on the BabyAI environment, covering tasks of varying complexity, and demonstrate positive transfer across agents. Additionally, we examine the planner’s generalisation ability to unseen agents and compare our method to traditional imitation learning approaches. By training on a pooled dataset from multiple agents, our universal policy achieves an improvement of up to $42.20\%$ in task completion accuracy compared to a policy trained on a dataset from a single agent\footnote{\url{https://github.com/NikeHop/UniversalPolicies}}.
\end{abstract}


\newcommand{\BibTeX}{\rm B\kern-.05em{\sc i\kern-.025em b}\kern-.08em\TeX}

\begin{document}


\pagestyle{fancy}
\fancyhead{}


\maketitle 


\section{Introduction}

 The challenge of developing a generalist agent capable of addressing a diverse range of sequential decision-making tasks remains an open problem \cite{gato}. Successfully solving this challenge holds the potential to eliminate the need for task-specific engineering and retraining, while also enhancing performance through positive transfer between tasks \cite{mt_transfer,octo_policy}. Given the vast differences in observation and action spaces across tasks, most general agents are still developed for specific domains such as robotic manipulation \cite{octo_policy,gr_2}, web agents \cite{mind_agent}, computer control \cite{os_control_general_agent}, or embodied navigation \cite{general_navigation}. However, image-based observations offer a common ground for many sequential decision-making tasks, even when the corresponding action spaces vary significantly. Image observations are prevalent in contexts such as gameplay \cite{atari_dataset,minecraft_dataset}, robotic control \cite{open_e_dataset}, and web or computer interfaces \cite{mind_agent,android_environment}, among others. Training a policy that can solve tasks that have environment observations in a unified format, such as images, is an important step towards creating a generalist agent. In this work, we move closer to this goal by developing a policy capable of controlling agents that share a common observation space but have different action spaces.

Recently, universal policies \cite{universal_policies} have emerged as a promising framework for learning multi-task policies through text guided video generation. This method is composed of two stages: first, a conditional diffusion model is trained to translate task descriptions into image-observation sequences; then, an inverse dynamics model is applied to map these sequences to the corresponding actions. A key advantage of this approach is the ability to pretrain the video generation model on vast datasets of instruction-video pairs \cite{howto100,ego4d,laion} or directly leveraging text-to-video foundation models \cite{imagen_video,sora_world_model1} as the planner. Additionally, the visual nature of the generated plans enhances interpretability by allowing inspection of the proposed solutions. Another potential advantage, which has not been investigated yet, lies in creating a shared planner for multiple agents, which can be paired with agent-specific inverse dynamics models to produce a policy capable of controlling all agents.

We explore this problem in a cross-agent setting where a group of agents shares the same observation space but operates with different action spaces. For each agent, there exists a small dataset of instruction-trajectory pairs. However, a single agent-specific dataset alone does not contain enough demonstrations such that training an agent-specific instruction-following policy through imitation learning on top of it would yield a policy capable of adequately solving the tasks for this agent. Our objective is to develop a policy that can successfully solve tasks for all agent types by pooling data from individual agents. To achieve this, we extend the universal policy framework to learn a policy on the combined dataset capable of controlling all the different agents.

The primary challenge lies in ensuring that the diffusion-based planner accounts for the varying capabilities of each agent (see Figure \ref{fig:example_plans_babyai}); otherwise, it may generate observation sequences that cannot be labelled via the agent-specific inverse dynamics model, leading to unpredictable behavior. On the other hand, this approach offers the potential for positive transfer, as the planner is exposed to a larger number of examples during training, which could lead to improved policies for each agent type. We investigate different methods for conditioning the planner on agent type information and assess their ability to generalise to unseen agents. In summary, our contributions are as follows:

\begin{figure}
    \centering
    \begin{minipage}{0.49\linewidth}
        \centering
        \includegraphics[width=\textwidth]{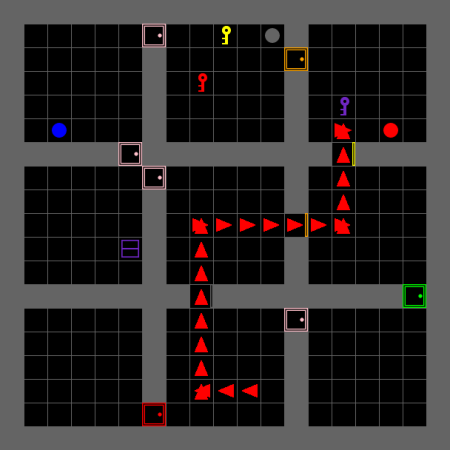}
    \end{minipage}
    \hfill
    \begin{minipage}{0.49\linewidth}
        \centering
        \includegraphics[width=\textwidth]{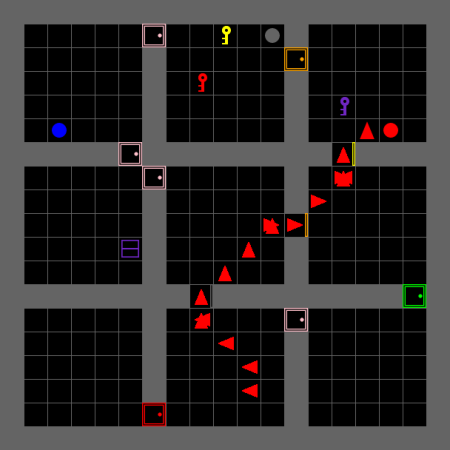}
    \end{minipage}
    \caption{Example for the different plans the shared planner needs to generate for different agent types. The agent on the left follows the standard action space (forward, turn left, turn right) in the BabyAI environment while the agent on the right can move to any of the surrounding squares and turn right.}
    \label{fig:example_plans_babyai}
\end{figure}
 
\begin{enumerate}
    \item Extending the universal policy framework to train a policy on a dataset consisting of trajectories from a diverse set of agents, capable of controlling all of them. 
    \item Comparing different approaches to condition the shared planner on agent type information and studying their ability to generalise to unseen agents.
\end{enumerate}

\section{Related Work}
We begin with an overview of approaches to training a generalist agent—a policy capable of solving a wide range of sequential decision-making tasks. Next, we discuss algorithms developed for cross-embodiment policy learning (CEPL) as the data setup is similar to ours. Finally, we review work focused on using diffusion models to learn control policies.

\subsection{Building generalist agents}

The GATO agent \cite{gato} represents the policy trained across the most diverse set of sequential decision-making tasks. It accomplishes this by mapping various action and observation spaces into tokens, which are then processed and predicted by a transformer architecture \cite{attention}. The effectiveness of this approach relies on the presence of generalizable patterns within the tokenized trajectory data, enabling positive transfer. A bottleneck in training general policies is the need to train policies from scratch which is computationally intensive. To address this, researchers have begun integrating multi-modal foundation models \cite{gpt_4,palm_e,pali_x} into their policies \cite{say_can,fm_seq_decision_making,rt2}. An example is the RT-2 model \cite{rt2}, which translates robotic actions into tokens and co-finetunes PaLI-X \cite{pali_x} and Palm-E \cite{palm_e} vision-language models using robotic trajectory data. Similarly, our adaptation of the universal policy approach can be combined with text-to-video foundation models \cite{sora_world_model1,imagen_video}.

\subsection{Cross Embodiment Policy Learning}

In robotics where data collection is costly, the community created increasingly larger datasets of task-trajectory pairs \cite{language_table,robonet,droid,open_e_dataset}. One possibility to scale the amount of available data quickly is to pool data over robotic agents with different embodiments. For example the Open X-Embodiment (OXE) dataset pools several existing robotic manipulation datasets \cite{open_e_dataset}. Learning over different embodiments comes with the challenge of learning a policy that needs to be able to handle different action and observation spaces.

While training agent-agnostic policies is not a new goal \cite{agent_agnostic_policy}, the embodiment gap has widened significantly, such that robot behaviour is even learned from human demonstrations \cite{xskill,learning_human_data1}. To address this challenge, two main approaches have emerged. The first maps different observation and action spaces of various embodiments into token sequences with a uniform structure, allowing them to be processed by a shared transformer \cite{octo_policy,cross_former}. The second focuses on learning a joint latent space that aligns different embodiments within a common embedding space, enabling policy learning over this latent representation \cite{xskill,latent_space_alignment}.

Our own data setup mirrors the OXE dataset as it pools agent-specific datasets from different agents. However, in the case of the OXE dataset, agents have varying observation spaces in addition to their differing action spaces. To learn a policy that can control a diverse set of agents we follow the universal policy approach~\cite{universal_policies} and frame policy learning as learning a generative model for observation sequences in combination with agent-specific inverse dynamics models. We further show that training a shared diffusion planner achieves greater positive transfer than imitation learning approaches with separate policy heads for each agent \cite{octo_policy,cross_former}.




\subsection{Policy Learning with Diffusion}

The recent success of diffusion models in learning generative models for images \cite{ldm,imagen}, audio \cite{audio_diffusion}, and videos \cite{imagen_video} has extended to policy learning \cite{diffuser,diffusion_policy,3d_diffusion_policy,universal_policies,seer}. These models can learn policies by either mapping states to action distributions \cite{diffusion_policy,3d_diffusion_policy} or modelling probability distributions over agent trajectories as sequences of state-action pairs \cite{diffuser}. In the latter approach, a trajectory is sampled based on a given task description and initial observation and then executed by the agent. Another approach proposed by \citet{universal_policies}, known as universal policies, learns a generative model over observation sequences conditioned on natural language instructions. Instead of generating action sequences directly, it generates observation sequences, which are then labelled with appropriate actions using an inverse dynamics model. This method not only outperforms imitation learning baselines \cite{rt1} and trajectory-based diffusion models \cite{diffuser} but also shows potential for integration with text-to-video foundation models. Additionally, it enhances interpretability by allowing inspection of the generated observation sequences. If agents share a common observation space, the diffusion-based planner for universal policies can be trained using trajectories from agents with varying capabilities. However, at test time, the generated plans must conform to the specific agent’s capability constraints. To the best of our knowledge, we are the first to explore diffusion-based generative modeling of observation sequences on datasets containing trajectories from agents with heterogeneous action spaces.

\section{Methodology}
We begin by extending the theoretical framework underlying universal policies called Unified Predictive Decision Process (UDPD) \cite{universal_policies} to be compatible with the cross agent setting. Then we introduce the data setup assumed for our approach, followed by a description how we implement the shared diffusion planner that can be conditioned on different types of agent information.

\subsection{CA-UDPD}

A cross agent UDPD (CA-UDPD) is a tuple $G = (N,X,C,H,p)$, where $N$ is a set of agents, $X$ represents the observations space for all agents, $C$ is a space of task descriptions, $H$ the time horizon and $p(\cdot |x_{0},c,k): X \times C \times N \rightarrow \Delta (X^{H})$ a conditional probability distribution over H-step observation sequences that depends on the current observation $x_{0}$, the task $c$ and the agent $k$ executing the task. Here $\Delta(Z)$ denotes a probability distribution over the space $Z$. The set of task descriptions as well as the observation space can take different forms. For example $C$ can be the set of all natural language instructions and $X$ the set of all RGB-images. The UDPD framework was introduced to address common challenges encountered when modelling real-world problems with MDPs \cite{mdp} and to shift focus on observation generation as a solution technique to learning policies. Instead of using a reward function, UDPD employs a more general set of task descriptions, recognizing that reward functions are often difficult to define in a way that consistently leads to the desired behaviour \cite{goal_misgeneralization}. Additionally, the action space and transition dynamics are not explicitly modelled but are captured within the conditional probability distribution over observation sequences. For a more detailed comparison between MDPs and UDPDs, we refer readers to the original paper \cite{universal_policies}. For control an agent-specific policy $\pi_{k}: X^{H} \rightarrow \Delta(A_{k}^{H-1})$ that maps observation sequences to a probability distribution over a sequence of actions from the agents action space $A_{k}$ needs to be learned for every agent type $k$.

\subsection{Problem Setup}\label{subsec:problem_setup}

For each agent $ n \in N $, we have a dataset $ D_{n} $ consisting of $ M_{n} $ instruction-trajectory pairs, denoted as $ D_{n} = \{(c_{i}, x_{1:t_{i}}, a_{i:t_{i}})\}_{i=1}^{M_{n}} $, where $ c_{i} \in C $ represents the instruction for the $i$-th sample and $a_{i:t_{i}}$ represents the action sequence the agent chose to complete the task, generating the observation sequence $x_{1:t_{i}} \in X^{t_{i}}$. Regarding the relation of the observation space of the different agents we consider two extreme cases. In the first case, all agents share the same observation space, i.e., $ X_{n} = X_{m} = X  \quad \forall  n, m \in N $. In the second case, the observation spaces are entirely disjoint, such that $ X_{n} \cap X_{m} = \emptyset \quad  \forall  n, m \in N $. In the latter scenario, the agent identifier is inherently embedded in the observation $ x_{t} $, allowing the planner $p$ to identify the agent from the observation $x_{0}$ it is conditioned on and plan accordingly. Both scenarios are plausible, i.e. for example in the case of robotic manipulation \cite{open_e_dataset} the image from the end-effector cameras would not let the planner identify which agent to plan for, while the static cameras will often show the agent the planner controls. We pool the individual datasets $D_{n}$ to obtain a larger mixture dataset $D=\{(c_{i}, x_{1:t_{i}}, a_{i:t_{i}},n_{i})\}_{i=1}^{M}$, where $n_{i} \in N$ is the agent id of the $i$-th sample and $M=\sum_{n=1}^{N} M_{n}$ is the total number of trajectories aggregated over all datasets. 

\subsection{Universal Cross Agent Policy}\label{subsec:planner}

\begin{figure*}[h!]
    \centering
    \includegraphics[width=\linewidth]{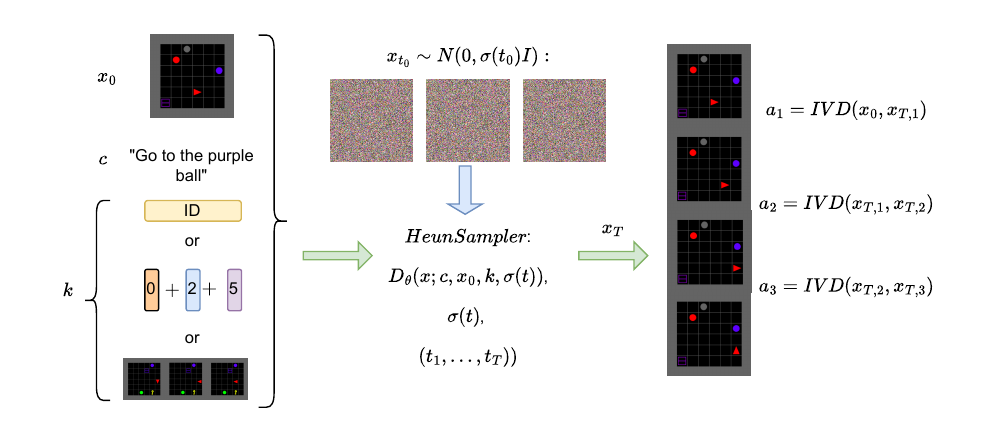}
    \caption{Overview of how actions are generated by UCAP. Given the starting observation $x_{0}$, the instruction $c$ and the different types of agent information $k$ to condition on, a Heun Sampler is applied for T steps to generate an observation sequence of 3 timesteps from noise. Two consecutive observations are then labelled via the inverse dynamics model to produce the action sequence the agent will take.}
    \label{fig:method_overview}
\end{figure*}

The goal is to train the conditional observation sequence generator $p(\cdot|x_{0}, c, k)$ on the mixture dataset $D$, leading to a \textbf{U}niversal \textbf{C}ross \textbf{A}gent \textbf{P}olicy (UCAP). Instead of generating the complete observation sequence $x_{1:t_{i}}$ we sample random windows of size 4 and take the first timestep to be the starting observation $x_{0}$, i.e. $p(\cdot|x_{0}, c, k)$ plans the next three timesteps for agent $k$ following instruction $c$ starting in $x_{0}$. In contrast to previous work we do not apply temporal super resolution, i.e first generating a coarse video plan that is refined in a second step \cite{universal_policies}.  We estimate $p$ using a conditional diffusion model. Diffusion models perturb data samples by gradually adding noise at different scales, then learn to reverse this process to recover the original data. New samples are generated by starting with noise and iteratively applying the learned reverse process to approximate the data distribution \cite{ddpm,song}. 

We follow the ODE formulation from \citet{elucidating_design_space}. Let $p_{\textrm{data}}(x)$ be the data distribution and $p(x;\sigma)$ be the perturbed data distribution obtained by adding i.i.d Gaussian noise with standard deviation $\sigma$ to it, i.e. $p(\tilde{x};\sigma)=\int p_{\textrm{data}}(x)p_{\sigma}(\tilde{x}|x)dx$, where $p_{\sigma}(\tilde{x}|x) = N(\tilde{x}|x,\sigma)$ \cite{score_based_generative_models_sde}. Given a continuous time noise schedule $\sigma(t)$, the solution to the following probabilistic flow ODE:
\begin{equation}
    dx = - \dot{\sigma}(t) \sigma(t) \nabla_{x} \log p(x;\sigma(t)) dt
\end{equation}
has marginals corresponding to the noise perturbed data distribution with noise $\sigma(t)$, i.e $x_{t} \sim p(x_{t},\sigma(t))$ \cite{elucidating_design_space}. If we have an estimate of the time dependent score function $\nabla_{x} \log p(x;\sigma(t))$ we can sample from the noise distribution $x_{t_{0}} \sim N(0,\sigma(t_{0})\textrm{I})$ and use ODE solvers \cite{ode_book} to turn the noise into samples from the data distribution. Following \citet{elucidating_design_space} this can be done by estimating a denoising function $D_{\theta}$ for different noise scales $\sigma$ via the loss:
\begin{equation}
    L(\theta) = \mathbb{E}_{x \sim p_{\textrm{data}}} \mathbb{E}_{n \sim N(0,\sigma I)} \|D_{\theta}(x+n;\sigma) -x \|^{2}_{2},
\end{equation}
where the denoising function $D_{\theta}$ and the score function are related via $\nabla_{x} \log p(x;\sigma) = (D(x;\sigma)-x)/\sigma^{2}$. At test time we then apply a sampler based on Heun's method \cite{ode_book,elucidating_design_space} to generate observation sequences from noise.

Following \cite{elucidating_design_space}, we condition the denoising network $D_{\theta}$, on the conditioning information, i.e. the starting observation $x_{0}$, the natural language instruction $c$ and the agent information $k$ to train a conditional diffusion model. The natural language instruction is embedded via a variant of the T5 model \cite{t5_encoder} and then added to the noise vector. Following \citet{universal_policies} we condition on the starting observation by concatenating it along the channel dimension for each timestep of the noise perturbed input. We could not find any benefit from classifier-free guidance \cite{cfg} and set the guidance weight to zero. Next we describe different ways to condition the denoising network $D_{\theta}$ on agent information.

\textbf{Agent ID:} For each agent type a random embedding serves as the agent id and is added to the noise embedding in the same manner as the instruction embedding. For example in case of robotic manipulation \cite{open_e_dataset} the planner would be conditioned on an embedding for each robot type (Franka Emika Panda, Sawyer, xArm, etc.). Since there is no relation between the ID and the capabilities of the agent, this type of conditioning cannot generalise to unseen agents with unseen IDs. 

\textbf{Action Space Representation:} If the action space is discrete and partly shared, one can represent the action space via a binary vector $v \in \{0,1
\}^{|A|}$, where $A$ is the union of all action agent-specific action spaces. If $v_{i}=1$ the agent is capable of action $i$. To condition the denoising network $D$ on this action space representation, we embed $v$ via a linear layer and add it to the noise embedding. This way of conditioning can potentially generalise to new agent types that have unseen action combinations, as the actions themselves have previously been encountered. 

\textbf{Example Trajectory:} Similar to the ability of LLM's \cite{gpt_4,mistral} to perform in-context learning \cite{in_context_learning1,in_context_learning2}, we can condition the planner on example observation sequences of the agent acting. 
If trained with a large enough number of different agent types and examples, it potentially generalizes to unseen agents by providing an example video of the novel agent's capabilities. However, taking a random subsequence of frames from a video of the agent acting as context is unlikely to be informative about the agent. For example in the BabyAI environment \cite{babyai} shown in Figure \ref{fig:example_plans_babyai}, sampling a subsequence where an agent only moves forward is uninformative of whether the agent can move diagonally or not. Conditioning on longer example videos increases the likelihood of the random video to be informative but also increase computational cost. In case of a discrete action space one option is to demonstrate each action the agent is capable of once. We condition on the resulting example video by concatenating it to the front of the noise perturbed sample along the time dimension.

Given a planner that can generate observation sequences, a set of inverse dynamics models needs to be trained for each agent type to label the sequences with the corresponding actions. An inverse dynamics model for agent $ n \in N$ is a function that takes two consecutive observations and maps them into the agent-specific action space $\mathrm{IVD}_{n}: X \times X \rightarrow A_{n}$. Given an observation sequence that is labelled with actions we can sample consecutive observation pairs and train the IVD to predict the correct action via the cross entropy loss. Since inverse dynamics models are sample efficient to learn they can be easily trained on the smaller agent specific dataset $D_{n}$. An overview of how a universal policy selects its actions can be seen in Figure \ref{fig:method_overview}. The details of the network architectures and training parameters for the conditional diffusion planner as well as the inverse dynamics models can be found in Appendix \ref{app:training_details}.

\section{Experiments}

We begin with outlining the experimental environment that facilitates the creation of agents with varying action spaces. Next, we investigate whether training on a pooled dataset, composed of trajectories from agents of different types, results in positive transfer. Additionally, we examine the impact of conditioning the planner on the different types of agent information presented in Section \ref{subsec:planner}. In the sections that follow, we compare the performance of UCAP against imitation learning baselines tailored to our data setup and conclude with ablations that investigate the performance of UCAP in the case of disjoint observation spaces, the effect of planning granularity on performance and the effect of number of agent types on generalisation performance.

\subsection{Environment}

\begin{figure}[h]
    \centering
    \begin{minipage}{0.23\textwidth}
        \centering
        \includegraphics[width=\textwidth]{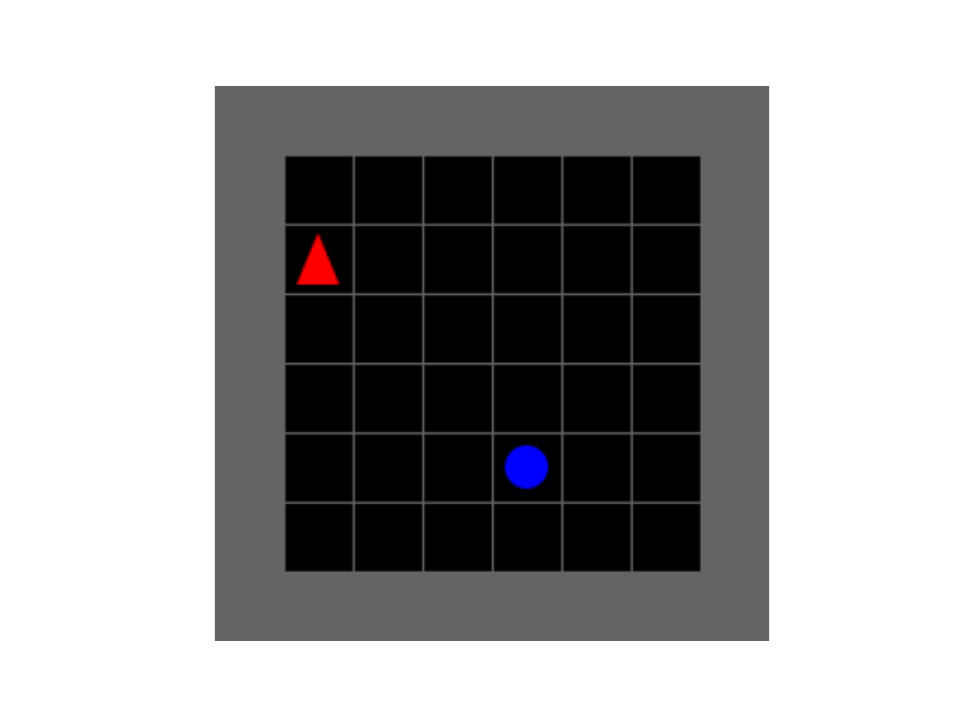}
        \caption*{Go to the blue ball.}
    \end{minipage}
    \hfill
    \begin{minipage}{0.23\textwidth}
        \centering
        \includegraphics[width=\textwidth]{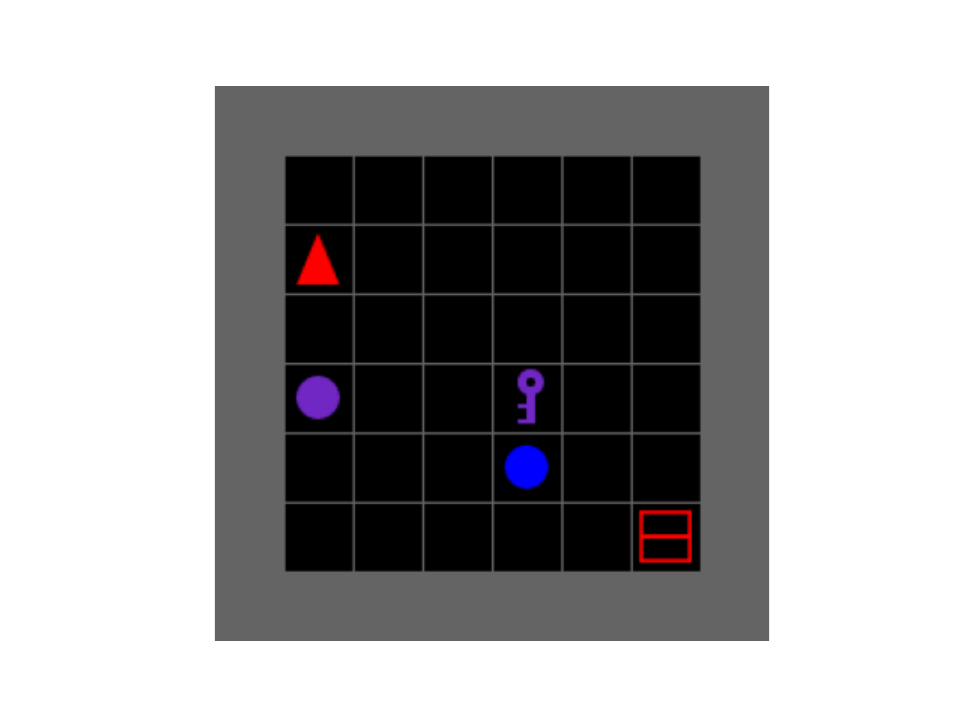}
        \caption*{Go to the purple key.}

    \end{minipage}
    \hfill
    \begin{minipage}{0.48\textwidth}
        \centering
        \includegraphics[width=\textwidth]{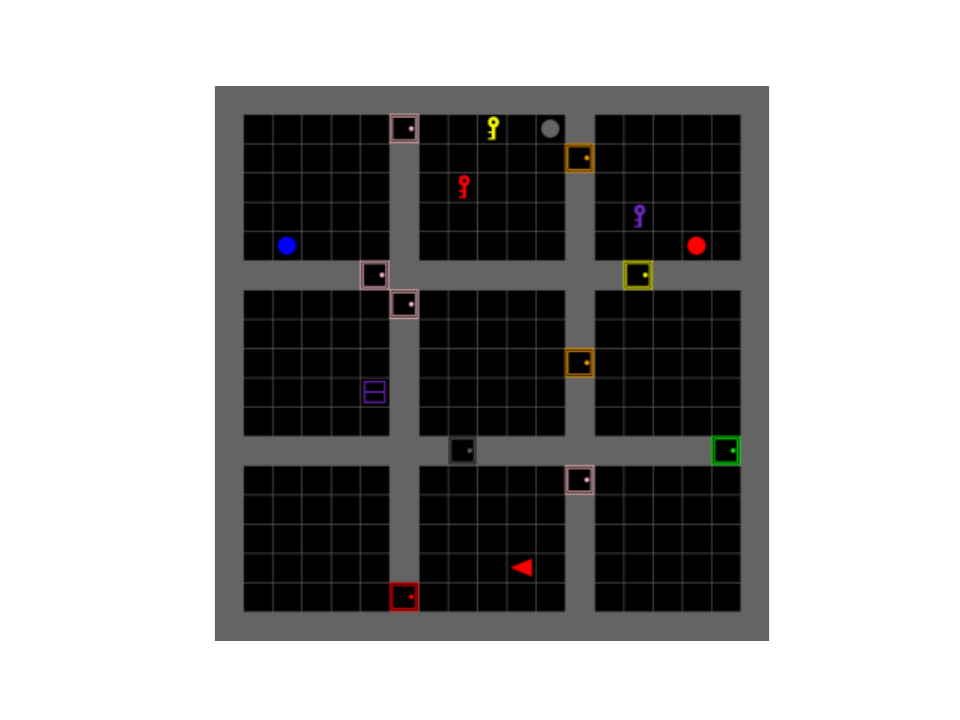}
        \caption*{Go to the red ball.}
    \end{minipage}
    \caption{Example environment observations from GoToObj (top left), GoToDistractor (top right) and GoToDistractorLarge (bottom). The agent is the red triangle.}
    \label{fig:environment_observation}
\end{figure}
We choose BabyAI \cite{babyai} as the evaluation environment since it is easily modifiable and offers a range of tasks with varying complexity. In the BabyAI environment an agent needs to navigate a gridworld to complete different tasks ranging from navigating to objects to opening doors with keys. The environment state can either be represented as a gridworld or an RGB image. Here we choose the gridworld representation as it is computationally more efficient. The environment state is fully observable. An agent's performance is evaluated by the percentage of tasks completed correctly over 512 episodes.
The standard action space consists of six actions (turn left, turn right, move forward, pickup, drop, toggle). We extend the action space to a total of 18 actions (e.g. turn 180 degrees, move diagonal, etc). To create different action spaces for different agents we mask out actions that are not available for the specific agent. In total we create eight agent types. An overview of the different agent types and their capabilities can be found in Table \ref{tab:action_spaces}. A summary of all the possible actions can be found in Appendix \ref{app:environment}. The mixture dataset $D$ is created by pooling the demonstrations of agents 0-5 (see Table \ref{tab:action_spaces}). The other two agents are left out to test generalisation of models to unseen agents. We refer to agents whose datasets are included in the pooled dataset as In-Distribution (ID) agents, while agents whose datasets are not included are called Out-of-Distribution (OOD) agents. We test the universal policy approach on the following three environment instances \footnote{https://github.com/Farama-Foundation/Minigrid}: 

\textbf{GoToObj:} The agent needs to navigate to the only object in the environment. The object can be a key, box, or ball, and comes in one of six different colours. This task does not require any natural language task instructions as the task is always the same. The environment observations have size $8 \times 8 \times 3$ and the trajectory length is at most 25 timesteps. 

\textbf{GoToDistractor:} The same environment as the GoTo environment but with 3 distractor objects added. Now the natural language instruction is necessary to understand to which object to navigate.

\textbf{GoToDistractorLarge:} The goal of the environment is the same as in the GoToDistractor environment. However the agent is now in one of nine rooms that are connected via doors and the environment contains 7 distractor objects. The environment observation is of size $22 \times 22 \times 3$ and the maximum trajectory length is at most 100 timesteps. An image representation of the different environment observations can be seen in Figure 
\ref{fig:environment_observation}.

\begin{table}[]
    \centering
    \caption{Overview of the different action spaces. The mixture of datasets used for training is derived from the ID agent types 0-5 and the OOD agent types are 6-7.}
    \begin{tabular}{l|l}
      AS   & Description \\
      \midrule
       0  & \makecell[l]{Standard: All actions of the standard action space \\ are allowed.} \\ 
       1  & No left: The agent cannot turn left. \\
       2  & No right: The agent cannot turn right. \\
       3  & \makecell[l]{Diagonal: The agent can additionally move \\ a diagonal step forward either to the right or left.} \\
       4 & \makecell[l]{WSAD: The agent can move one field to the right left, \\ up and down  and turns into the direction it moves.}\\
       5 & \makecell[l]{Dir8: The agent can move to any \\ of the eight surrounding fields and turn right.} \\
       6 & Left-Right: The agent can go left and right and turn right.  \\
       7 & \makecell[l]{All Diagonal: The agent can go to all the diagonal cells\\ and turn right.} \\
    \end{tabular}
    \label{tab:action_spaces}
\end{table}

\subsection{Positive Transfer}

\begin{figure*}[t!]
    \centering
    \begin{minipage}{0.32\textwidth}
        \centering
        \includegraphics[width=\textwidth]{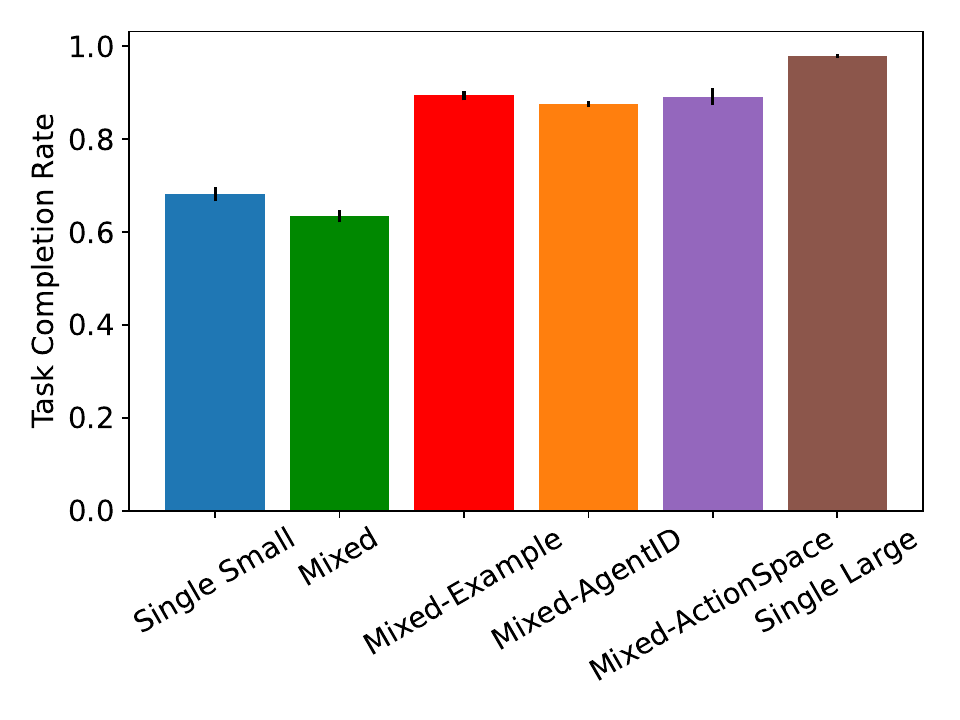}
    \end{minipage}
    \hfill
    \begin{minipage}{0.32\textwidth}
        \centering
        \includegraphics[width=\textwidth]{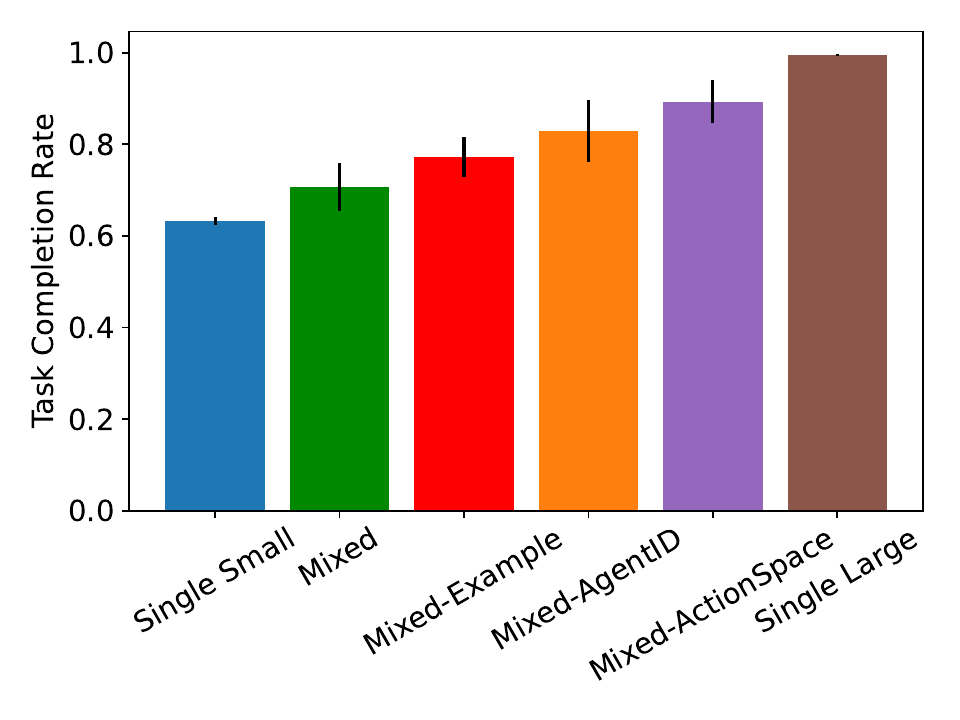}

    \end{minipage}
    \hfill
    \begin{minipage}{0.32\textwidth}
        \centering
        \includegraphics[width=\textwidth]{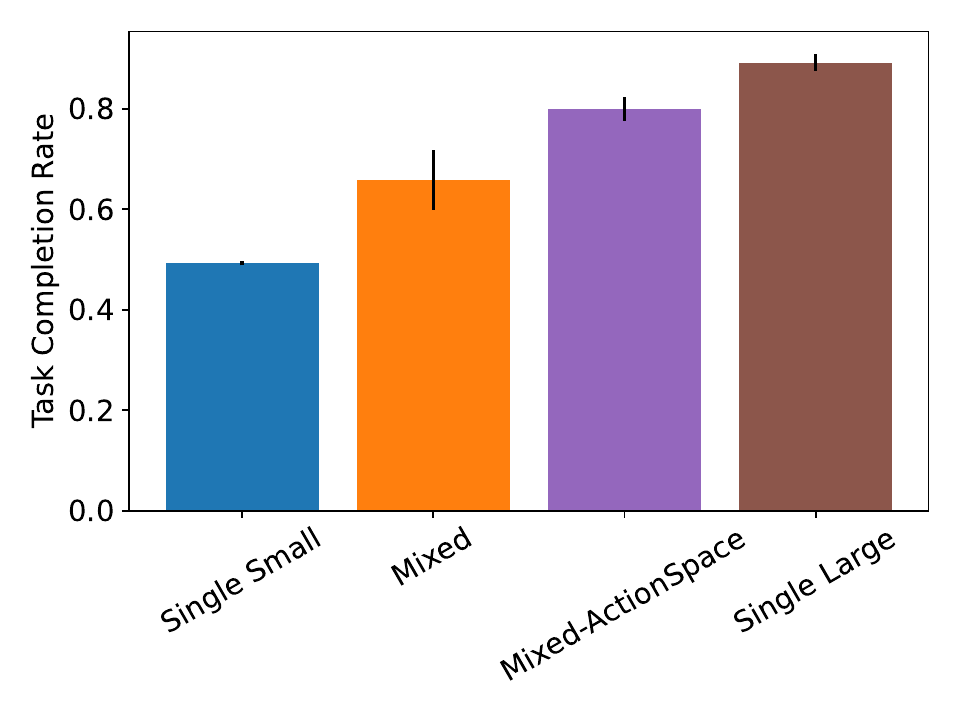}
    \end{minipage}
    \caption{Mean task completion rate for a range of UCAP models on the three evaluation environments GoToObj (left), GoToDistractor (middle), GoToDistractorLarge (right) for an agent with the standard action space which belongs to the ID agent set. All results are averaged over 4 random seeds and error bars indicate the standard error.}
    \label{fig:transfer_results}
    
\end{figure*}

Positive transfer occurs if UCAP outperforms on average the universal policies trained for a specific agent type on the agent specific dataset.
In order to investigate whether positive transfer occurs we perform the following steps:

\begin{enumerate}
    \item Train a universal policy for a single agent on a small agent-specific dataset for each agent type. 
    \item Train a universal policy on the mixture of small datasets (UCAP). Here we train a policy for each type of agent information from Section \ref{subsec:planner} (Mixture-Example, Mixture-AgentID, Mixture-ActionSpace) and one UCAP that is not conditioned on any agent information (Mixture).
    \item Train a universal policy for a each agent type on an agent-specific dataset equal in size to the mixture dataset.
\end{enumerate}

The performances of the universal policies trained on the small and larger dataset serves as anchor to evaluate the amount of positive/negative transfer that occurs. The policy generated by training on the large dataset is an upper bound to the amount of transfer we can expect. Any performance improvement above the policy trained on the smaller dataset is a sign of positive transfer, and any performance decrease is a sign of negative transfer. For the two smaller environments we perform the analysis for all agent types. For the GoToDistractorLarge environment, we only do the analysis for the standard action space, i.e. no single agent policies are trained for the other agent types to save computational resources.

In Figure \ref{fig:transfer_results} we can see the performance of the different universal policies in the three environments for the standard agent. To save compute we only train UCAP conditioned on the action space encoding for the GoToDistractorLarge environment since it worked best in the smaller environments. For all environments UCAP conditioned on any type of the agent representation shows positive transfer, where conditioning on a representation of the action space works best and conditioning on examples has the worst mean task completion rate. One possible reason for the weaker performance when conditioning on examples is their higher dimensionality compared to agent ID and action space encodings, such that it takes more samples to  estimate the conditional probability distribution correctly. The results for the diffusion planner without agent information are mixed showing positive transfer in the more complex environment and negative transfer in the simplest environment. One potential reason is that for more complex environments the agent will encounter unseen distractor goal combinations if not enough training data is present. Having observed these combinations even in trajectories from other agents can be beneficial for the planning performance. Therefore it can be beneficial to train on the mixture dataset even if no agent information is present. However, conditioning on agent information offers large performance benefits.

Table \ref{tab:results_transfer} shows the task completion rates of UCAP conditioned on different action space information, averaged across agent types, for both ID and OOD agents. For ID agents, the results mirror those from the standard action space, except that the planner without agent conditioning does not exhibit negative transfer in the GoToObj environment. In GoToObj, transfer effects depend on the agent type: more general action spaces allow for better execution of planner-proposed actions, leading to positive transfer. For instance, in GoToObj, the mean task completion rate drops from $0.827 \pm 0.023$ to $0.617 \pm 0.016$ for agents that cannot turn right but increases from $0.705 \pm 0.058$ to $0.731 \pm 0.026$ for agents that can move diagonally. For OOD agents, UCAP conditioned on examples or action space information shows no generalization. A possible reason is the limited agent diversity, causing conditioned information to be treated as ID. The planner may struggle to correlate actions in video examples with those in demonstrations due to insufficient variability in agent types. This hypothesis will be further examined in an ablation study in subsection \ref{subsec:ablation}.

\begin{table}[]
    \centering
    \caption{Mean task completion rate averaged over all ID and OOD agents respectively for different UCAP models in the GoToObj environment and GoToDistractor environment. All results are averaged over 4 seeds and brackets contain standard errors. The best performing models for ID agents as well as OOD agents are in bold, where models trained on large individual datasets are excluded.}
    \begin{tabular}{l|c|c}
        Model & \multicolumn{2}{c}{GoToObj-Env} \\
         & ID Agents & OOD Agents \\
        \midrule
        Single Agent Small & $0.753 (0.009)$ & $\mathbf{0.657 (0.005)}$ \\
        Mixed & $0.760 (0.003)$ & $ 0.396 (0.014)$ \\
        Mixed Example & $\mathbf{0.857 (0.005)}$ & $0.453 (0.010)$ \\
        Mixed Agent-ID & $0.850 (0.004) $ & $0.261 (0.055) $ \\
        Mixed ActionSpace & $0.850 (0.009)$ & $0.494 (0.021)$ \\
        Single Agent Large & $0.915 (0.002)$ & $0.970 (0.004)$ \\
        \midrule
         & \multicolumn{2}{c}{GoToDistractor-Env} \\
         & ID Agents & OOD Agents \\
        \midrule
        Single Agent Small & $0.627 (0.003)$  & $\mathbf{0.650 (0.007)}$  \\
        Mixed & $0.726 (0.047)$ & $0.315 (0.017)$  \\
        Mixed Example & $0.795 (0.042) $ & $0.404 (0.024)$  \\
        Mixed Agent-ID & $0.843 (0.060)$ & $0.193 (0.019)$  \\
        Mixed ActionSpace & $\mathbf{0.892 (0.053)}$ & $0.541 (0.034)$ \\
        Single Agent Large & $0.995 (0.001)$ & $0.996 (0.002)$ \\

    \end{tabular}
    \label{tab:results_transfer}
\end{table}

\subsection{Baselines}

\begin{table*}[t!]
    \centering
    \caption{Average task completion rate averaged over all ID agents and OOD agents respectively for imitation learning (IL) baselines in comparison to universal policies (UP) trained on single agent datasets and mixture datasets. Results are averaged across four random seeds and standard errors are in brackets. The naming of the model consists of first the method name, then the dataset trained on and thirdly an indicator whether finetuning is performed after training. The best performing method that has access to only the small individual agent datasets is in bold.}
    \begin{tabular}{l|c|c|c|c}
       Model  & \multicolumn{2}{c}{GoToObj-Env} & \multicolumn{2}{c}{GoToDistractor-Env} \\
       & ID Agents & OOD Agents & ID Agents & OOD Agents \\
       \midrule
       IL - Single Agent Small & $0.638 (0.010) $ & $0.390 (0.017)$ & $0.504 (0.006)$ & $0.514 (0.018)$ \\
       IL Union of Action Spaces - Mixture & $0.763 (0.022)$ & $0.030 (0.004)$ & $0.812(0.005)$ & $0.026 (0.002)$ \\  
       IL Union of Action Spaces - Mixture - Finetuned & $\mathbf{0.880 (0.003)}$ & $0.705 (0.009) $ &  $0.803 (0.029)$ &  $0.7028 (0.031)$ \\
       IL Agent Heads - Mixture  & $0.818 (0.008)$ & $0.012 (0.001)$ & $0.801 (0.018)$ & $0.016 (0.005) $ \\ 
       IL Agent Heads - Mixture - Finetuned & $0.827 (0.003)$ & $0.771 (0.008)$ & $0.811 (0.037)$ & $0.742 (0.044)$ \\
       UCAP - Action Space  & $0.850 (0.009)$ & $0.494 (0.021)$ & $\mathbf{0.892 (0.053)}$ & $0.541 (0.034)$ \\
       UCAP - Action Space - Finetuned & $0.839 (0.007)$  & $\mathbf{0.788 (0.011)}$ & $0.872 (0.046)$ & $ \mathbf{0.904 (0.039)}$ \\
       IL - Single Agent Large & $0.907 (0.003)$ & $0.955 (0.009)$ & $0.953 (0.006)$ & $0.944 (0.001)$ \\
    \end{tabular}
    \label{tab:imitation_learning_results}
\end{table*}

We compare UCAP to imitation learning baselines adapted to learning from multiple smaller datasets from different agents. The policy architecture for all imitation learning baselines consists of a convolutional stack followed by an MLP \cite{baby_ai_architecture}. Policies are trained to predict the agent's next action given a state sampled from the expert demonstrations. More details on the architecture and hyperparameters can be found in Appendix \ref{app:imitation_learning}. We implement the following approaches:

\textbf{Imitation Learning (IL)}: We train a single agent policy on the small and large dataset. Again the performance on the large dataset serves as an upper bound of imitation learning, while the performance on the smaller dataset serves as a threshold to whether we observe positive or negative transfer. 

\textbf{IL - Union of Action Spaces}: A common way of handling different actions spaces is to form the union over all agents action spaces \cite{heterogenous_ma_rl}. The model receives a one-hot encoded vector representing the agent ID, which is concatenated with the output of the convolutional embedding stack. If the agent chooses an invalid action there is no change to the environment state. 

\textbf{IL - Agent Heads}: Following the approach of the Octo-policy \cite{octo_policy} we employ different agent heads but keep the same convolutional backbone, i.e. for each agent a separate MLP is trained to predict the action the agent of the specific type needs to take. 

We implement finetuned versions of the imitation learning baselines, where the policy is first trained on the large mixture dataset and then finetuned on smaller agent-specific datasets. For comparison, we also finetune a universal policy without agent information after training it on the mixture dataset.

Table \ref{tab:imitation_learning_results} shows the imitation learning baselines compared to the universal policy conditioned on an action space encoding. Both imitation learning variants show positive transfer, with task completion rates improving when trained on the mixture dataset instead of the small single-agent dataset in both environments. As expected, without finetuning, the imitation learning baselines cannot generalize to OOD agents. For policies with varying agent heads, a new agent requires predicting actions via an untrained head, while the imitation learning baseline receives an unseen agent ID as input. The universal policy conditioned on action space information outperforms both baselines in both environments, with a larger performance gap in the more complex environment. None of the methods generalize to OOD agents. The standard approach to handle this is finetuning on a small dataset from the unseen agent \cite{octo_policy}. The finetuned universal policy outperforms the finetuned imitation learning baselines, with the gap widening in more complex environments. This suggests it is easier to finetune the observation sequence planner than to learn a new agent head or adapt a large action space distribution to a new one-hot encoding.

\subsection{Ablations}\label{subsec:ablation}

\subsubsection{Disjoint Observation Spaces}

\begin{figure}
    \centering
    \includegraphics[width=\linewidth]{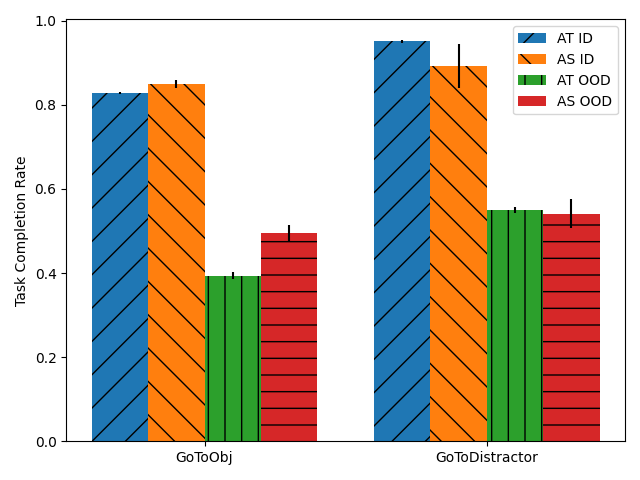}
    \caption{Mean task completion rate over all ID and OOD agent types in the GoToObj and GoToDistractor environment for the universal policy trained on a mixture of dataset with a visual agent type (AT) and without a visual agent-type but conditioned on an encoding of the action space (AS). Results are averaged over 4 random seeds and error bars indicate standard errors.}
    \label{fig:agent_id_ablation}
\end{figure}

As mentioned in Section \ref{subsec:problem_setup} the observation space of different agents does not necessarily need to overlap, i.e. the observation itself contains information on the agent type. The planner could potentially leverage this information to adjust the plan of the agent accordingly. To emulate this setting in the BabyAI environment, we give each agent type a different colour and train the universal policy on a newly created mixed dataset where agents are identifiable via their colour. In Figure \ref{fig:agent_id_ablation} we compare the results to the best UCAP model with uniformly coloured agents in the GoToObj and GoToObjDistractor environment. The planner successfully leverages the colour information and reaches a task completion rate comparable or higher than the UCAP conditioned on an action space encoding. As expected the planner does not generalise to OOD agents with an OOD colour. 

\subsubsection{Planning Granularity}
An alternative strategy to handle differing action spaces is to plan at a coarser timestep, where the planner suggests the next goal state, and each agent’s goal-conditioned policy chooses the action to move from the current state to the proposed goal. This approach has the benefit that the generated plans do not need to conform with the agents capabilities at a timestep level, but transfers the responsibility from the planner to the agent-specific goal-conditioned policies. Compared to learning an inverse dynamics model, developing a local goal-conditioned policy typically requires more data due to the increased complexity of the mapping. Here, local refers to the fact that the goal is reachable within a few timesteps. 

\begin{figure}
    \centering
    \includegraphics[width=\linewidth]{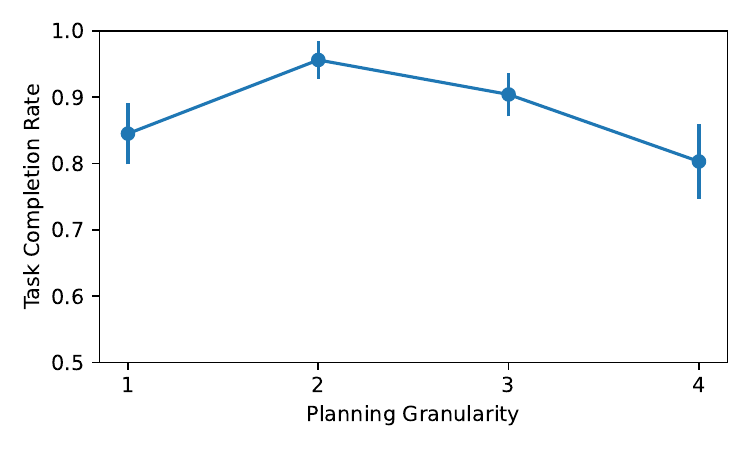}
    \caption{Task completion rate for diffusion planners with different planning granularities trained on the mixture dataset.}
    \label{fig:abl_planning_granularity}
\end{figure}

To test this, we trained the diffusion planner on subsampled trajectories in the GoToDistractor environment, selecting the start and end states along with every $n$-th timestep, and trained a goal-conditioned policy through imitation learning by sampling start and goal states from the agent-specific datasets. First, we evaluated only the planner’s performance across different planning granularity levels using an oracle policy, which moved the agent to each planner-suggested goal state. In Figure \ref{fig:abl_planning_granularity} we can see that the best planning performance is reached when planning for two timesteps. Evaluating the 2-step planner with the agent-specific goal-conditioned policies shows also an improved task completions rate compared to combining a 1-step planner with agent-specific IVD models (see Table \ref{tab:abl_planning_granularity}). Notably, the performance on the OOD agents is on par with the performance for the ID agents as the 2-step planner cannot propose impossible actions for agents anymore.

\begin{table}[]
    \centering
    \begin{tabular}{l|c|c}
         Model  & \multicolumn{2}{c}{GoToDistractor-Env} \\
        & ID Agents &  OOD Agents \\
        \midrule
        UCAP 1-Step  & $0.726(0.024)$ & $0.315(0.09)$ \\
        UCAP 2-Step  & $\mathbf{0.885 (0.011)}$ & $\mathbf{0.918 (0.010)}$ \\ 
    \end{tabular}
    \caption{Average task completion rate for UCAP with different planning granularity trained on the mixture dataset. Experiments are run over 4 seeds and numbers in brackets indicate standard error.}
    \label{tab:abl_planning_granularity}
\end{table}

\subsubsection{Increased Agent Diversity}

One possible reason why conditioning on either an encoding of the action space or trajectory examples fails to generalize to OOD agents is the limited diversity of agents in the training set. With only six different agents, it may be difficult to learn the correlation between conditioning information and the possible actions in the trajectory to be reconstructed. To address this, we created a dataset for the GoToDistractor environment, consisting of trajectories from 6,595 different agents by combining the 18 possible actions in various ways. Notably, the dataset size remains unchanged compared to the previous setup. As shown in Table \ref{tab:gen_ability}, this increased diversity improves the generalisation performance of both conditioning methods. The mean task completion rate increases from $0.494\pm0.021$ to $0.906\pm0.010$ when conditioned on the action space encoding and from $0.453\pm0.10$ to $0.611\pm0.01$2 when conditioned on example trajectories.

\begin{table}[]
    \centering
    \begin{tabular}{l|c}
         Model  & OOD Agents \\
        \midrule
        UCAP Action Space &  $0.906(0.010)$ \\
        UCAP Example &  $0.611(0.012)$ \\ 
    \end{tabular}
    \caption{Average task completion rate for UCAP trained on the mixed dataset with different conditioning information for OOD agents. Experiments are run over 4 seeds and numbers in brackets indicate standard error.}
    \label{tab:gen_ability}
\end{table}

\section{Limitations}

While we observed positive transfer from training on a pooled dataset of different agents, it is unclear whether this transfer will hold when environment observations vary more significantly, such as in the OXE dataset \cite{open_e_dataset}, which includes observations from diverse robotic manipulators. Future work should attempt to scale the approach to larger datasets with heterogenous A drawback of using the diffusion planner is the increased training and inference time. Convergence requires approximately 40 times more updates compared to imitation learning, and the sampler needs 128 neural function evaluations versus just one for imitation learning. Techniques like progressive distillation \cite{progressive_distillation} and consistency models \cite{consistency_models} improve sampling speed with minimal loss in generative ability.

\section{Conclusion}
We showed that it is possible to leverage the universal policy approach \cite{universal_policies} to train a diffusion based planner that generates observations sequences for agents with different capabilities. Training the planner on a pooled dataset from all agent types leads to an improved universal policy compared to training on the smaller agent specific datasets alone. The performance of the planner can be improved by conditioning on agent specific information such as the encoding of the action space or example observation sequences. Independent of the agent information the planner is conditioned on generalisation to OOD agents without finetuning remains a challenging task, where potential solution strategies include increased agent diversity and planning at a higher granularity level. Future work should extend the approach to agents with heterogenous observation spaces \cite{octo_policy,cross_former} and scale it to more complex environments~\cite{open_e_dataset}. 

\newpage

\begin{acks}
This research was (partially) funded by the Hybrid Intelligence Center, a 10-year programme funded by the Dutch Ministry of Education, Culture and Science through the Netherlands Organisation for Scientific Research, https://hybrid-intelligence-centre.nl. This work used the Dutch national e-infrastructure with the support of the
SURF Cooperative using grant no. EINF-6630.
\end{acks}

\bibliographystyle{ACM-Reference-Format} 
\bibliography{sample}

\clearpage
\appendix

\section{BabyAI Environment}\label{app:environment}

We adapted the official BabyAI implementation \footnote{https://github.com/Farama-Foundation/Minigrid} to handle a larger range of actions. The complete list of actions can be found in Table \ref{tab:action_space}. We used the BabyAI bot to generate the demonstrations for the different agent types. In Table \ref{tab:sample_size} one can find the the different dataset sizes used for each environment instance.

\begin{table}[bp]
    \centering
    \caption{The extended action space for the BabyAI environment.}
    \begin{tabular}{c|l}
       Action ID  &  Effect \\
       \midrule
       0  &  Turn left \\
       1  &  Turn right \\
       2  &  Move forward \\
       3  &  Pick up \\
       4  &  Drop \\
       5  &  Toggle \\
       6  &  Done \\
       7  &  Diagonal Left \\
       8  &  Diagonal Right \\
       9  &  Move Right \\
       10 &  Move Down \\
       11 &  Move Left \\
       12 &  Move Up \\
       13 &  Turn Around \\
       14 &  Left Diagonal Backwards \\
       15 &  Right Diagonal Backwards \\
       16 &  Move Left no turn \\
       17 &  Move Right no turn \\
       18 &  Backward \\
    \end{tabular}
    \label{tab:action_space}
\end{table}

\begin{table}[bp]
    \centering
    \caption{Dataset sizes for a single agent dataset that is used for creating the pooled dataset for all three environments. Note that this refers to the number of trajectories contained in the dataset. For the GoToDistractorLarge (GoToDist.-Large) environment the trajectories consists of more timesteps. As a result more instruction-subsequence pairs can be sampled from a single trajectory in the larger environment in comparison to a trajectory from the smaller environments.}
    \begin{tabular}{l|c|c|c}
       Dataset  & GoToObj & GoToDist. & GoToDist.-Large \\
       \midrule
       Single Agent  & 1494 &  90000 & 25000 \\
    \end{tabular}
    \label{tab:sample_size}
\end{table}

\section{Training details}\label{app:training_details}

\subsection{Conditional Diffusion Model} \label{app:diffusion_model}

We modify the implementation of \citet{elucidating_design_space} \footnote{https://github.com/NVlabs/edm} to handle the conditioning on the different agent informations presented in Section \ref{subsec:planner}. We adopt their hyperparameters of the diffusion process (see Table \ref{tab:hyperparameter_diffusion_process}) and keep them constant across all three environments (GoToObj, GoToDistractor, GoToDistractorLarge). For the number of sampling steps we choose $64$ as increasing the number of sampling steps did not lead to performance improvements worth the additional computational effort. The training specific hyperparameters for all three environments can be found in Table \ref{tab:training_hyperparameters}.

\begin{table}[bp]
    \centering
    \caption{The hyperparameters of the conditional diffusion model adopted from \citet{elucidating_design_space}. These values were used for all three environment instances.}
    \begin{tabular}{l|c}
      Parameter & Value \\
      \midrule
      $\sigma_{min}$ & 0.002 \\
      $\sigma_{max}$ & 80 \\
      $\sigma_{data}$ & 0.5 \\
      $\rho$ & 7  \\
      $P_{\textrm{mean}}$ & -1.2 \\
      $P_{\textrm{std}}$ & 1.2 \\
      sampling steps & 64 \\
    \end{tabular}
    \label{tab:hyperparameter_diffusion_process}
\end{table}

\begin{table}[bp]
    \centering
    \caption{The training hyperparameters for all three environments. The Hardware used for training the model in the GoToDist. environment was a NVIDIA A100 partitioned into two instances with 20GB each.}
    \begin{tabular}{c|c|c|c}
       Parameter  & GoToObj & GoToDist. & GoToDist.-Large \\
     \midrule
       \# Updates & 500.000 & 1.000.000 & 500.000 \\
       Learning Rate & 0.00002 & 0.00002 & 0.00005\\
       Batch-Size & 64 & 128 & 512 \\
       Hardware & \makecell{NVIDIA \\ TITAN X \\ (Pascal) \\  (12GB)} & \makecell{NVIDIA \\ -A100 \\ (40GB)-MIG}  & \makecell{NVIDIA \\ -A100 \\ (40GB)} \\
       
    \end{tabular}
    \label{tab:training_hyperparameters}
\end{table}

\subsection{Inverse Dynamics Model}
The inverse dynamics model for each agent are trained on the small agent-specific datasets. The input observations are concatenated along the channel dimension and processed by a convolutional block with residual connections. Then mean pooling is applied across all pixels and the resulting embedding is processed via a linear layer. The inverse dynamics model is trained via the cross-entropy loss. The training parameters can be found in Table \ref{tab:training_parameters_ivd}.

\begin{table}[bp]
    \centering
    \caption{Training parameters for the inverse dynamics models. In case of multiple values, they refer to the values used for the GoToObj, GoToDistractor and GoToDistractorLarge environment respectively.}
    \begin{tabular}{l|c}
         Parameter & Value  \\
         \midrule
         \# Epochs &  500/10/100\\
         Learning Rate & 0.0001 \\
         Batch-Size & 64 / 64 / 256   \\
         Hardware & \makecell{NVIDIA TITAN X (Pascal)(12GB)}
    \end{tabular}
    \label{tab:training_parameters_ivd}
\end{table}

\subsection{Imitation Learning}\label{app:imitation_learning}

We follow the architecture from \citet{baby_ai_architecture}, but remove the LSTM-component since we perform imitation learning in the fully observable variant of the BabyAI environment. The training details for all imitation learning baselines can be found in Table \ref{tab:training_details_imitation_learning}. All models were trained until convergence of the action prediction accuracy on the validation dataset.

\begin{table}[bp]
    \centering
    \caption{Training parameters for the different imitation learning baselines for the GoToObj and GoToDistractor environment. In case of multiple values they refer to the methods Imitation Learning - Small Single Agent Dataset, Imitation Learning - Large Single Agent Dataset, Imitation Learning - Unione of Action Spaces and Imitation Learning - Agent Heads respectively. Single values indicate that the value is shared across all baselines.}
    \begin{tabular}{l|c|c}
        Parameter & GoToObj & GoToDist \\
        \midrule
        \# Epochs & 100/25/100/25  & 50/20/30/20 \\
        Learning Rate & 0.0001 & 0.0001 \\
        Batch-Size & 64 & 128 \\
        Hardware & \makecell{NVIDIA\\ TITAN X (Pascal)\\(12GB)} & \makecell{NVIDIA\\ TITAN X (Pascal)\\(12GB)}\\
    \end{tabular} \label{tab:training_details_imitation_learning}
\end{table}

\end{document}